\documentclass[journal]{vgtc}      
\onlineid{0}

\vgtccategory{Research}

\vgtcpapertype{application}

\title{LatentFlow: Visual Analytics for Latent Space Analysis in \\ Molecular Graph Neural Networks}

\author{
  \authororcid{Shiyi Liu}{0009-0009-0174-0952},
  \authororcid{Jiaqing Chen}{0009-0002-9395-0277},
  \authororcid{Nicholas Hadler}{0000-0002-9167-5227},
  \authororcid{Rostyslav Hnatyshyn}{0009-0006-0510-1152},
  \authororcid{Michael W. Mahoney}{0000-0001-7920-4652},\\
  \authororcid{Talita Perciano}{0000-0002-2388-1803},
  \authororcid{John F. Hartwig}{0000-0002-4157-468X},
  \authororcid{Gunther H. Weber}{0000-0002-1794-1398},
  \authororcid{Ross Maciejewski}{0000-0001-8803-6355}
}

\authorfooter{
\item 
    Shiyi Liu, Jiaqing Chen, Rostyslav Hnatyshyn, and Ross Maciejewski are with Arizona State University.
    E-mail: shiyiliu@asu.edu; jchen501@asu.edu; rhnatysh@asu.edu, and rmacieje@asu.edu; 
\item
    Nicholas Hadler is with University of California, Berkeley.
    E-mail: nhadler@berkeley.edu
\item 
    Michael W. Mahoney is with ICSI, LBNL, and University of California, Berkeley.
    E-mail: mmahoney@stat.berkeley.edu
\item 
    Talita Perciano and Gunther Weber are with Lawrence Berkeley National Laboratory.
    E-mail: TPerciano@lbl.gov and GHWeber@lbl.gov
\item
    John F. Hartwig is with University of California, Berkeley, and Lawrence Berkeley National Laboratory.
    E-mail: jhartwig@berkeley.edu.
}

\abstract{%
  Chemists and materials scientists increasingly use machine learning models, such as graph neural networks (GNNs), to predict properties of molecules and the outcomes of their reactions.
  Beyond predictive performance, understanding how these models organize chemical information internally in their latent spaces, i.e., the embeddings of the molecules, is critical. 
  Analyzing latent spaces helps diagnose model behavior and assess whether the learned embeddings are organized in ways that reflect meaningful chemical relationships.
  Unfortunately, existing methods provide limited support for analyzing latent spaces across layers and across different model states  (e.g., training epochs, model configurations, and input data), making it difficult to understand how these latent spaces evolve throughout a model or relate to chemical concepts.
  We present \emph{LatentFlow}, a visual analytics system developed in collaboration with a domain expert for analyzing latent spaces in molecular GNNs. 
  \emph{LatentFlow} groups embeddings into clusters and supports exploration of latent spaces by tracking how these clusters change across layers and model states using a modified Sankey diagram.
  To support interpretation, \emph{LatentFlow} links these clusters to representative molecules and their shared substructures, and it allows scientists to introduce their own domain knowledge and compare it with the patterns found in the latent spaces.
  We evaluate \emph{LatentFlow} through two case studies. 
  The results show that \emph{LatentFlow} helps scientists understand how latent spaces evolve, identify meaningful molecular patterns, and better interpret model behavior.
}


\teaser{
  \centering
  \includegraphics[width=\linewidth]{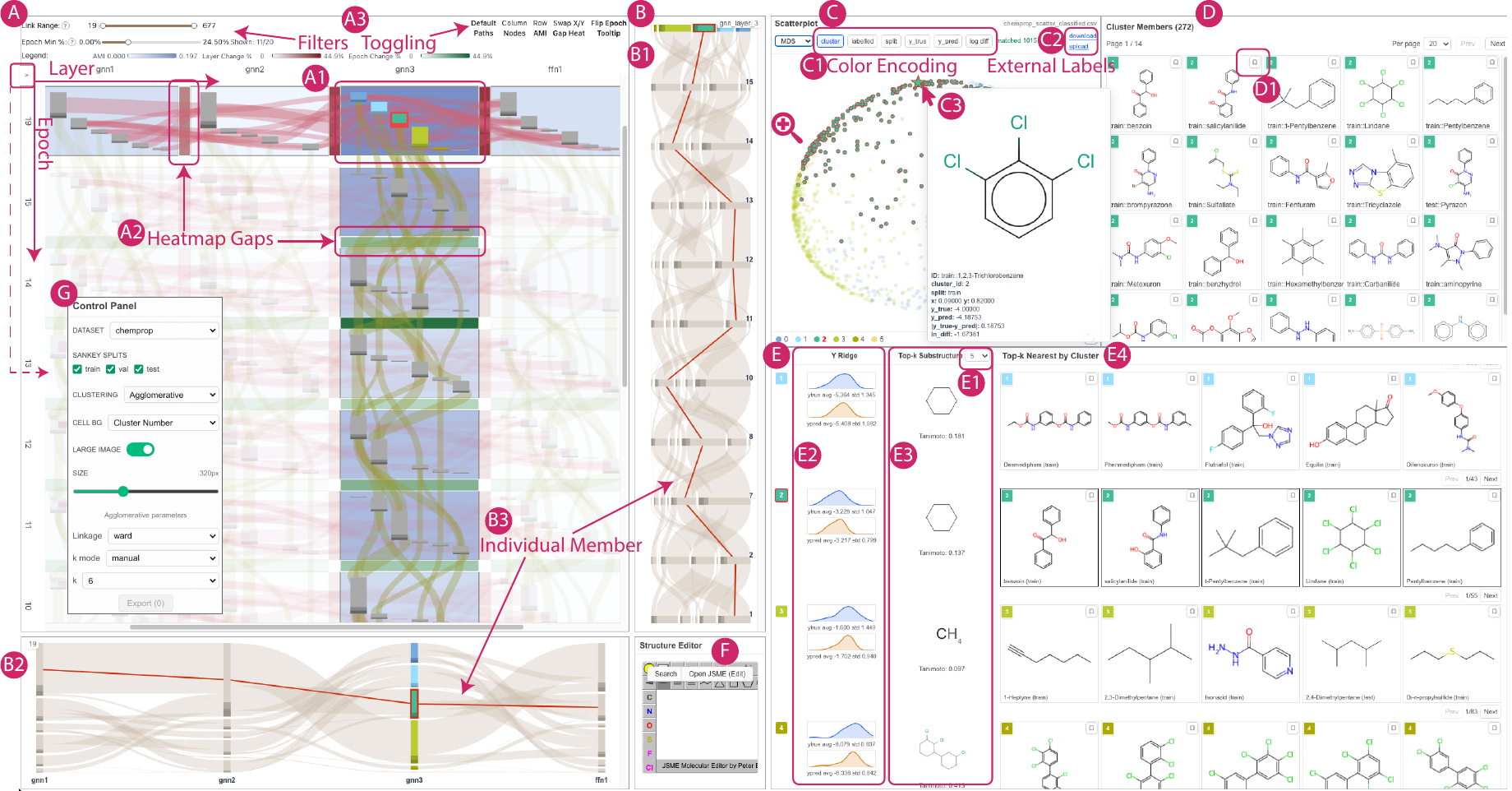}
  \caption{
    \emph{LatentFlow} interface for analyzing latent spaces in molecular GNNs through coordinated views. 
    (A) The Modified Sankey View shows cluster changes across two model-state dimensions. 
    (B) The horizontal (B1) and vertical Sankey View (B2) reveal detailed transitions along a single dimension. 
    (C) The scatterplot projects embeddings into 2D to inspect how molecules are distributed and separated. 
    (D) The cluster member view displays molecules within a selected cluster for direct inspection. 
    (E) The Top-$k$ Panel summarizes cluster-level structure, including the $k$ value selector (E1), prediction distributions (E2), substructures (E3), and representative molecules (E4).
    (F) The Structure Editor allows experts to query substructures and highlight matching molecules across views. 
    (G) The control panel provides controls for dataset selection, clustering configuration, and coordinated interaction.
    }
  \label{fig:system}
}

\graphicspath{{figs/}{figures/}{pictures/}{images/}{./}} % where to search for the images

\usepackage{booktabs}                  % only used for the table example
\usepackage{mathptmx}                  % use matching math font
\usepackage{amssymb}

\usepackage{amsmath}
\usepackage{graphicx}
\usepackage{color}
\begin{document}

%%%%%%%%%%%%%%%%%%%%%%%%%%%%%%%%%%%%%%%%%%%%%%%%%%%%%%%%%%%%%%%%
%%%%%%%%%%%%%%%%%%%%%% START OF THE PAPER %%%%%%%%%%%%%%%%%%%%%%
%%%%%%%%%%%%%%%%%%%%%%%%%%%%%%%%%%%%%%%%%%%%%%%%%%%%%%%%%%%%%%%%

%% The ``\maketitle'' command must be the first command after the
%% ``\begin{document}'' command. It prepares and prints the title block.
%% the only exception to this rule is the \firstsection command
% \firstsection{Introduction}

\maketitle
\clearpage
\section{Introduction}
In chemistry and materials science, machine learning models, especially Graph Neural Networks (GNNs), are widely used to study molecular properties and chemical reactions~\cite{atzGeometricDeepLearning2021a, reiserGraphNeuralNetworks2022,shiReviewApplicationsGraph2024}.
By operating directly on graph representations of molecular structure, GNNs have achieved state-of-the-art performance in tasks such as molecular property prediction~\cite{gasteigerDirectionalMessagePassing2022b,chemprop}, reaction outcome prediction~\cite{Shilpa2023}, toxicity screening~\cite{Miao2023, MONEM2025109614}, and materials discovery~\cite{CHEN2023108506}.
Despite this success, molecular machine learning models often provide limited insight into why particular predictions are made.
Researchers, therefore, analyze the internal representations learned by GNNs to better understand how they encode and organize chemical information.

During learning, each layer transforms its input into a set of embeddings that collectively define a layer-specific \emph{latent space}.
These latent spaces are often projected into lower-dimensional spaces and clustered to reveal emergent patterns~\cite{andronovExploringChemicalReaction2021, burnsDeepLearningFoundation2026, bushuievSelfsupervisedLearningMolecular2025}.
However, it remains difficult to determine which structural patterns are represented at different layers, when chemically meaningful structure--property relationships emerge, and how these patterns evolve, reorganize, or deteriorate during training~\cite{JMLR:v19:17-646,li2022understandingcollapsenoncontrastivesiamese}.
Tracking latent-space evolution can therefore reveal how chemical information develops within the model and help experts identify the factors that ultimately drive its predictions.

We identify three key challenges in understanding how molecular representations are learned and organized in the latent spaces of molecular GNNs:
(i) tracing changes across multiple latent spaces; (ii) identifying meaningful patterns in large latent spaces; and (iii) relating latent spaces to chemical knowledge.
First, tracing changes across latent spaces is difficult because embeddings vary across \emph{layers} and \emph{model states}
~\cite{kornblith2019similarity}. 
\emph{Model states} refer to training epochs, model configurations, and input data.
Extant visualizations for tracing transitions operate along a single dimension, making it difficult to follow how molecules are reorganized across multiple dimensions.
Second, the scale of latent spaces makes it difficult to identify meaningful patterns, as a large number of points need to be manually examined. 
%Analyzing these spaces requires inspecting a large number of points, making it hard to locate relevant molecules and important changes.
Third, relating latent spaces to chemical knowledge remains challenging. Embeddings encode molecular structure in a numerical form that does not necessarily correspond to chemical concepts, obfuscating what drives the observed organization of molecules~\cite{jimenez2020drug, qadri2025explainable}.

% In this paper, we present \emph{LatentFlow} (\cref{fig:system}), a visual analytics system for analyzing how latent spaces evolve across layers and model states, and for identifying chemically meaningful patterns in molecular GNNs. 
In this paper, we present LatentFlow (\cref{fig:system}), a visual analytics system for exploring how latent spaces in molecular GNNs evolve across network layers and model states. LatentFlow helps computational chemists identify chemically meaningful patterns encoded in learned molecular representations.
The system was developed in close collaboration with a domain expert and follows a multi-level visual analytics workflow that progresses from overview to detailed inspection.
\emph{LatentFlow} groups molecules based on their embeddings in each latent space.
To support tracking how these clusters change, we introduce an interactive modified Sankey diagram that reveals the transitions between clusters of embeddings as they split, merge, or persist across layers and model states.
Interactive filtering and summary visualizations reduce visual clutter while preserving key structural patterns in cluster transitions.
% Beyond visualizing latent spaces, \emph{LatentFlow} connects model outputs to domain knowledge, enabling experts to interpret the results in terms of chemistry.
\emph{LatentFlow} connects model outputs to domain knowledge by linking clusters to molecular structures, substructures, and similarity measures, enabling experts to interpret the results in terms of chemistry.
Experts can also introduce their own domain knowledge by defining substructures or labels and compare them with model-learned patterns in the latent space. 
% In this way, \emph{LatentFlow} supports human-in-the-loop visual analytics for understanding how latent spaces form and evolve in chemical GNNs.
% \hfill \break
Our contributions are summarized as follows:
\begin{itemize}
    % \item We develop \emph{LatentFlow}, an interactive visual analytics system that supports the exploration of high-dimensional latent spaces down to individual molecules.
    \item We develop \emph{LatentFlow}, an interactive visual analytics system that enables chemists to explore high-dimensional latent spaces down to individual molecules.
    \item We propose a design for a modified Sankey diagram that enables experts to explore large latent spaces, discovering interesting molecular cluster patterns and tracing how these clusters split, merge, stabilize, or collapse.
    \item We validate our approach through two case studies, demonstrating \emph{LatentFlow}'s effectiveness in understanding model behavior and connecting model dynamics with chemically meaningful patterns.
\end{itemize}

\section{Related Work}
The design of \emph{LatentFlow} relates to the analysis of neural network representations, visual interpretability of machine learning models, and chemistry-specific visualization. 
In this section, we first review how researchers study latent spaces to understand the internal learning processes of deep neural networks (\cref{subsec:latentspace}); 
we then describe visual analytics systems developed to make these opaque models interpretable (\cref{subsec:valatent}); and, 
finally, we review visual analytics approaches applied to chemistry, positioning our work within the broader landscape of interpretable scientific machine learning and human-in-the-loop chemoinformatics (\cref{subsec:vachem}).

\subsection{Latent Spaces in Deep Neural Networks}
\label{subsec:latentspace}
Deep neural networks learn transformations that map raw inputs into latent spaces, i.e., high-dimensional embeddings that place similar data points close to each other~\cite{bengio2013representation, goodfellow2016deep}. 
These embeddings serve as input features for downstream tasks such as property prediction and classification, while also providing a basis for notions of similarity through distance.
%, and they also enable the analysis of similarity and grouping patterns among data points through clustering and distance-based methods.
Researchers often examine latent spaces to understand how features are learned and develop across layers~\cite{rauber2016visualizing}. 
%To support such analysis, prior work has proposed a range of methods to analyze and compare embeddings. 
A common approach uses similarity-based metrics, %to compare representations across layers, training epochs, or different models. 
which show that models may converge to structurally similar embeddings despite different initializations~\cite{raghu2017svcca, kornblith2019similarity}. 
%In addition, dimensionality reduction facilitates the inspection of 
Dimensionality reduction~\cite{hua2021feature} supports the inspection of embedding distributions and can reveal training and optimization issues such as embedding collapse, where embeddings lie in a lower-dimensional subspace, making distinct inputs difficult to distinguish.
Clustering and distance-based approaches characterize how data points organize in latent space. 
Embeddings often form structured groups aligned with semantic or functional patterns~\cite{bengio2013representation}.
Most existing methods still rely on aggregate similarity measures or static visualizations, making it difficult to examine how molecular groups form, evolve, and reorganize across layers and model states.

\subsection{Visual Analytics for Model Latent Spaces}
\label{subsec:valatent}
Beyond static projections, interactive approaches explore and analyze latent spaces. 
A number of visual analytics systems facilitate the exploration of latent spaces to help experts interpret and diagnose neural network behavior. 
To interpret these high-dimensional spaces, many prior approaches rely on dimensionality reduction to obtain low-dimensional representations of embeddings. Techniques such as PCA~\cite{MACKIEWICZ1993303}, t-SNE~\cite{van2008visualizing}, and UMAP~\cite{mcinnes2018umap} project these embeddings into two or three dimensions, where scatterplots remain the dominant visual form~\cite{huang2023embedding}.
These methods typically either emphasize the spatial arrangement of points in the projection~\cite{smilkov2016embedding} or relate the projection to known labels or properties~\cite{7192695,8440820}.
This allows experts to inspect the distribution of embeddings in the latent space, including cluster structure, separation, and density patterns; comparisons are supported by encoding cluster membership as color~\cite{7192695,8440820}.

Systems such as WizMap~\cite{wang-etal-2023-wizmap} further scale these scatterplot-based visualizations to embeddings with millions of data points.
EmbeddingVis~\cite{8802454} provides coordinated views to examine how relationships among data points are preserved in the embedding space. 
Latent space cartography~\cite{liu2019latent} enables experts to explore latent spaces by defining directions that correspond to meaningful changes in the data and compares how embeddings vary along these directions.
CorGIE~\cite{9705082} connects the original graph structure with the embedding space, allowing experts to see how each node is positioned in the latent space.

Despite these advances, prior work primarily analyzes embedding spaces in isolation or through pairwise comparison. 
To the best of our knowledge, none of the existing approaches explicitly models how representation structures change across model layers and model states. 
To address this gap, we introduce a modified Sankey diagram that represents the transition of molecules between clusters and supports visual analysis across two dimensions.

\subsection{Visual Analytics for Chemoinformatics}
\label{subsec:vachem}
Visualization has long played a central role in analyzing chemical data. 
% A common strategy is to project high-dimensional molecular fingerprints or learned embeddings into two- or three-dimensional space to assess structural similarity, diversity, and patterns.
Early work on visualizing learned chemical embeddings explored local sampling strategies, such as inspecting neighborhoods around exemplar molecules to assess embedding coherence~\cite{Bombarelli2018}. 
Extensions sample along principal components of pretrained embeddings~\cite{C8SC04175J} or traverse random orthogonal directions in latent space~\cite{Kusner2017}, providing qualitative insights into embedding smoothness and structural organization.

Building on projection-based analysis, numerous interactive systems have been developed to support analysis of chemical data. 
For example, InVADo~\cite{10334517} focuses on the interactive visual analysis of molecular docking data, supporting experts in examining docking poses, interaction patterns, and binding affinities through coordinated views tailored to structure-based drug discovery tasks. 
%Complementing such task-driven systems, 
Scalable projection techniques such as TMAP~\cite{Probst2020TMAP} enhance neighborhood preservation to enable the visualization of large molecular descriptor datasets. 
More comprehensive platforms, including PUMA~\cite{Medina2017} and DataWarrior~\cite{Thomas2015}, integrate multiple analytical techniques to support chemical diversity analysis, scaffold inspection, and interactive compound selection.

With the growing adoption of machine learning models in chemoinformatics, visualization has increasingly been paired with interpretability methods.
Systems such as the ChemInformatics Model Explorer (CIME)~\cite{Humer2022CIME} provide interfaces for inspecting datasets with visualizations of how individual atoms contribute to model predictions. 
ChemoGraph~\cite{Bharat2023} is an interactive visual analytics system that represents molecular similarity as a graph, allowing experts to explore groups of similar molecules. 
ChemVA~\cite{Sabando2021ChemVA} supports virtual screening of candidate chemical compounds through coordinated views that enable exploration of molecular similarity, comparison of multiple dimensionality reduction projections, and assessment of projection trustworthiness.
% and examine their local neighborhoods. 
% These approaches primarily focus on interpreting model outputs or feature importance at the molecule level, linking predictions to molecular substructures.
These approaches support the exploration of molecular similarity, projection structure, model outputs, or feature importance, but generally focus on individual molecular representations or predictions.
Our work introduces a visual analytics system that not only interprets model outputs at the molecular level, but also allows chemists to incorporate their own domain knowledge and compare it with the output of the model.

\section{Background}
Understanding the behavior of deep molecular models requires bridging machine learning mechanisms with chemistry. In this section, we outline the technical foundations that inform the visual analytics system of \emph{LatentFlow}. We first explain how molecular GNNs encode structural information to form dynamic latent spaces. Following this, we introduce the established chemical representations and similarity metrics that chemists rely on to interpret these molecular spaces.

\subsection{Molecular GNNs}
Molecular GNNs represent each molecule as a graph, where atoms form nodes and chemical bonds form edges. 
Node features encode basic chemical information, such as the atom's species, degree, aromaticity, formal charge, bond type, and other related structural properties. 
A molecular GNN associates each node (i.e., atom) with a vector called a node embedding. 
These embeddings are computed through message passing, where each node gathers information from its neighboring nodes and updates itself. 
Stacking multiple layers allows nodes to capture information from increasingly large neighborhoods. 

After message passing, the model can compute a single embedding for each molecule by combining node embeddings; for example, through summation or averaging.
Embeddings constantly change during training as model parameters are optimized across successive epochs. 
%These updates adjust how the model gathers information, leading to continuous shifts in the embeddings of the same molecules. 
Different training configurations further alter this process and produce different embeddings for the same set of molecules.

A \emph{latent space} consists of the embeddings of all molecules at a given layer in a model.
Each molecule corresponds to a point in this space, and its positions reflect how the model encodes structural similarity. 
Since embeddings change across both layers and model states, the latent space does not remain fixed but evolves over time.
As a result, the same set of molecules can be arranged differently across multiple latent spaces, each arrangement corresponding to a stage of training.

To examine how molecules are organized in a latent space, researchers commonly use clustering and dimensionality reduction. 
Clustering groups molecules based on their embeddings and assigns molecules with similar patterns to the same cluster, while dimensionality reduction projects high-dimensional embeddings into a low-dimensional subspace, providing a view of their overall distribution and arrangement.

\begin{figure*}
\centering
\includegraphics[width=\linewidth]{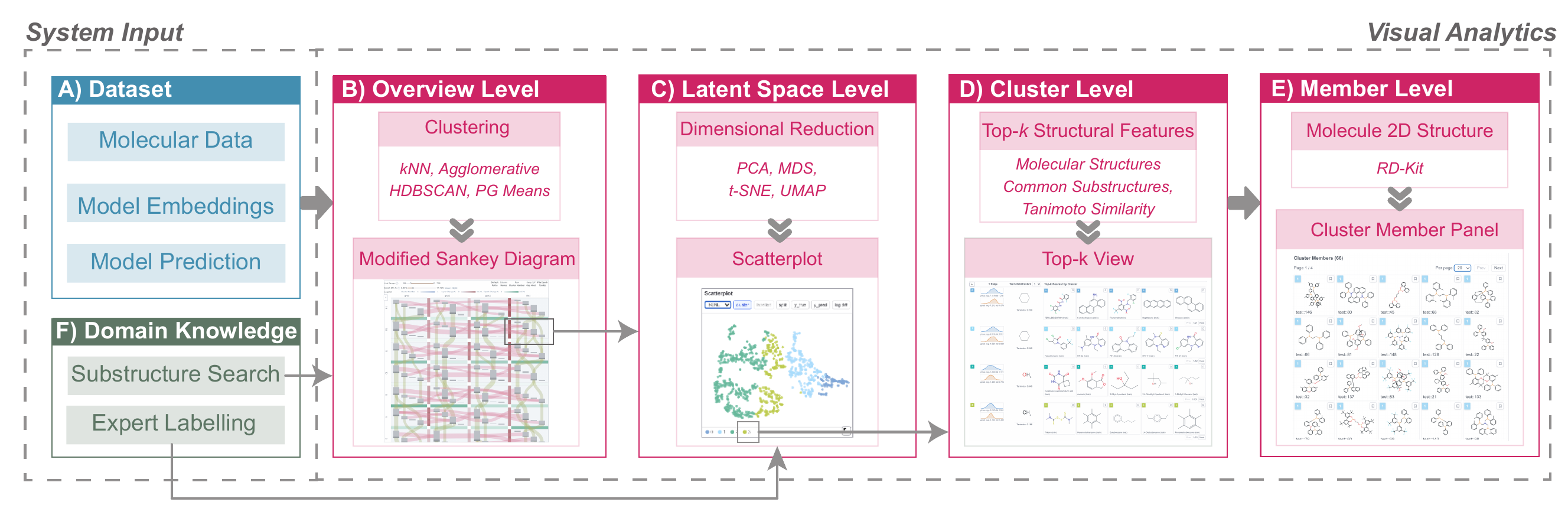}
\caption{
        \emph{LatentFlow} integrates model data and domain knowledge into a four-level visual analytics system. 
        (A) Model data, including molecular structures, embeddings extracted from different layers and model states, and model predictions, are provided as input. 
        (B) At the overview level, embeddings are grouped into clusters using clustering methods and visualized in the Modified Sankey View to show how molecules move between clusters. 
        (C) At the latent space level, embeddings are projected into two dimensions using dimensionality reduction methods and displayed in the Scatterplot View to reveal how molecules are distributed. 
        (D) At the cluster level, structural fingerprints are used to compute Tanimoto similarity and common substructures, which are shown in the Top-k View to summarize cluster structure. 
        (E) At the member level, individual molecules are rendered as 2D structures in the Cluster Member Panel, enabling detailed inspection and cross-view tracing. 
        (F) Domain knowledge is introduced through substructure queries and uploaded labels, which are applied across views to relate model-learned organization to known chemical patterns.
        }
    \label{fig:pipeline}
\end{figure*}

\subsection{Chemical Representations and Similarity}
\label{subsec:chembg}
Molecules are often compared through their substructures.
Small changes in chemical structure can lead to drastically different properties. 
For example, replacing a hydrogen atom with a chlorine atom in an aromatic ring can significantly decrease solubility and increase hydrophobicity.
For this reason, chemists often look for common substructures when they want to compare molecules.

Molecular structures are often represented using Simplified Molecular Input Line Entry System (SMILES) strings, which encode a molecule as a text sequence that can be converted into a 2D depiction.\footnote{\url{http://opensmiles.org/}}
This representation is convenient for storage, display, and retrieval.
%It also makes it possible to show the same molecule in both a compact textual form and a visual structure view.
To compare molecules, chemists often use structural fingerprints, which summarize the presence or absence of local patterns in a molecule. 
Tanimoto similarity~\cite{Bajusz2015} compares two fingerprints and returns a score that reflects how much their structures overlap. 
A higher score indicates that two molecules share more of the same structural features, while a lower score indicates the opposite.
Common substructures and similarity scores are useful because they help connect groups of molecules back to concrete chemical patterns. 

\section{Analytical Tasks and Requirements}
We developed \emph{LatentFlow} through an iterative design process in close collaboration with a domain expert in computational chemistry with approximately 7 years of experience. 
The collaboration lasted five months, during which we met twice per week to refine the system through continuous feedback and design iterations.
We initially began development by identifying three core challenges (\textbf{C1}--\textbf{C3}) that scientists typically encounter when they try to analyze the training dynamics of the models they are developing.
We subsequently derived four corresponding design requirements (\textbf{DR1}--\textbf{DR4}) from these challenges that inform our system's design.

\subsection{Challenges}
Understanding how molecular GNNs learn and organize chemical information is inherently challenging. 
Through our iterative design process with a domain expert, we identified three key challenges:

\noindent\textbf{C1. Tracking the evolution of latent spaces.}
Embeddings evolve across layers and model states, resulting in multiple latent spaces for the input molecules. 
Tracing how latent spaces organize molecules as embeddings change across these spaces is challenging because these variations occur along multiple dimensions.
Traditional flow-based visualizations (e.g., Sankey diagrams) can represent how groups evolve along one dimension, such as layers or training epochs, but they fail to capture relationships in multiple dimensions.
%However, they cannot capture changes arising from multiple factors at the same time, which does not match this multi-dimensional setting.

\noindent\textbf{C2. Identifying meaningful patterns.}
With multiple latent spaces derived from different layers and model states, the total number of embeddings grows quickly. 
Each latent space contains the entire input dataset, and collectively, they form a huge, high-dimensional dataset that is difficult to analyze directly.
While dimensionality reduction and clustering techniques provide summaries of individual latent spaces, they introduce a significant cognitive load, as the expert needs to inspect a large number of points and groups.
As a result, it is difficult to identify relevant molecules and important patterns.

\noindent\textbf{C3. Relating latent spaces to chemistry.}
%Embeddings define latent spaces and encode molecular structure in a numerical form. 
While embeddings capture patterns learned by the model, they do not necessarily correspond to established chemical concepts such as functional groups, bonding patterns, or other interpretable properties.
Even when molecules form clusters or are separated in the latent space, it remains unclear which aspects of molecular structure drive these patterns.

\subsection{Design Requirements}
We derived the following design requirements from the challenges identified above, which guided the development of \emph{LatentFlow}. 

\noindent\textbf{DR1. Multi-dimensional embedding evolution tracking.}
The system should provide a visualization that shows how molecules are reorganized across latent spaces from different layers and model states. 
The visualization should support at least two dimensions of variation, so that changes from different sources can be observed together.
This requirement addresses \textbf{C1}.

\noindent\textbf{DR2. Provide summaries and interactions for latent space exploration.}
The system should summarize each latent space to show how molecules are organized and how their embeddings evolve across each dimension, eliminating the need to perform manual pairwise comparisons of latent spaces. 
Experts should be able to filter latent spaces based on how much they change.
This requirement addresses \textbf{C2}.

\noindent\textbf{DR3. Link embeddings to molecules.}
The system should allow experts to identify which molecule each embedding represents and inspect it in detail. 
%Experts should be able to select an embedding and see the corresponding molecule, or select a molecule and locate its embedding in latent spaces. 
The same molecule should be highlighted across views to support exploration.
This requirement contributes to addressing \textbf{C3}.

\noindent\textbf{DR4. Show structural features for selected molecules.}
The system should be able to compute and display structural features for selected molecules. 
This feature would assist experts in understanding what structural properties are shared by these molecules and how similar they are. %, and whether their properties are consistent.
This requirement addresses \textbf{C2} and \textbf{C3}.

% \textbf{DR5. Support smooth interaction with large-scale data.}
% The system should handle large numbers of embeddings and latent spaces without noticeable delay, and support smooth interaction.

\section{LatentFlow}

% \subsection{System Overview}
\label{subsec:overview}
\emph{LatentFlow} supports the analysis of latent spaces by combining model data and domain knowledge through coordinated visualizations (\cref{fig:pipeline}). 
The system accepts two types of input: (1) model data, including molecular structures, embeddings extracted from different layers and model states, and model predictions; and (2) domain knowledge, which can be introduced by experts through substructure queries using the Structure Editor (\cref{fig:system}F, \cref{subsec:struc_editor}) or by uploading external labels (\cref{fig:system}C2, \cref{subsec:scatterplot}). 
These inputs are integrated through a multi-level visual analytics system that allows experts to move from an overview to a detailed inspection of individual molecules.

At the overview level (\cref{fig:pipeline}B), embeddings from each model state are grouped into clusters using methods such as k-nearest neighbors (k-NN)~\cite{1053964}, agglomerative hierarchical clustering (AHC)~\cite{Ward01031963, Johnson_1967}, hierarchical density-based spatial clustering (HDBSCAN)~\cite{hdbscan}, and projected Gaussian means (PG-Means)~\cite{pgmeans}. 
These methods produce different groupings: k-NN emphasizes local neighborhood structure, agglomerative clustering provides hierarchical groupings at different levels of granularity, HDBSCAN identifies clusters with varying densities and separates noise, and PG-Means adapts the number of clusters based on the data distribution. 
The Modified Sankey View (\cref{fig:system}A) visualizes the resulting clusters, representing clusters as nodes and transitions of molecules across layers and model states as flows (\textbf{DR1}). 
This view allows experts to identify regions where cluster structure changes significantly and select cells for further analysis (\textbf{DR1}).
When external labels are provided, similarity metrics such as adjusted mutual information (AMI)~\cite{MEILA2007873}, normalized mutual information (NMI)~\cite{1412045}, and the Fowlkes–Mallows index (FMI)~\cite{Fowlkes01091983} are computed to quantify the agreement between cluster assignments and the uploaded labels, 
%These metrics measure how consistently molecules grouped in the same cluster correspond to the same label category, 
with higher values indicating stronger agreement (\textbf{DR2, DR3}). 
The selected metric is encoded as the background color of each cell in the Modified Sankey View.

% At the latent space level (\cref{fig:pipeline}C), embeddings are projected into two dimensions using dimensionality reduction methods such as PCA, MDS, t-SNE, or UMAP. 
% These methods provide different views of the data: PCA emphasizes global variance, MDS preserves overall distance relationships, t-SNE highlights local neighborhoods and cluster separation, and UMAP balances local and global structure. 
% The resulting projections are shown in the Scatterplot View, where each point represents a molecule. 
% This allows experts to examine how molecules are distributed, assess cluster separation, and compare cluster structure with labels or model outputs (\textbf{DR3}).

At the latent space level (\cref{fig:pipeline}C), embeddings are projected into two dimensions using dimensionality reduction methods. 
To support comparison across model states, the projection pipeline employs temporally consistent layouts. 
PCA uses a shared projection basis across all model states, while t-SNE and UMAP employ temporally coherent layouts inspired by Dynamic t-SNE~\cite{tsne_joint} and Joint UMAP~\cite{10.1007/978-981-96-5815-2_14}.
These methods provide different views of the data: PCA emphasizes global variance, t-SNE highlights local neighborhoods and cluster separation, and UMAP balances local and global structure. 
The resulting projections are shown in the Scatterplot View, where each point represents a molecule. 
This approach allows experts to examine how molecules are distributed, assess cluster separation, and compare cluster structure with labels or model outputs (\textbf{DR3}).

At the cluster level (\cref{fig:pipeline}D), the system computes structural features to summarize each cluster (\textbf{DR4}). 
To compare molecules, structural fingerprints encode the presence of local chemical patterns, and their Tanimoto similarity score (c.f.~\cref{subsec:chembg}) is displayed.
Common substructures are extracted from representative molecules within each cluster. 
The Top-$k$ Panel displays these features, allowing experts to relate clusters to shared structural patterns and assess how consistent the molecular structure is within clusters.

At the member level (\cref{fig:pipeline}E), individual molecules are rendered as 2D structures using RDKit~\cite{landrum2016rdkit}. 
The Cluster Member Panel shows these representations, allowing experts to inspect molecules directly and compare structures (\textbf{DR3}). 
Interactions such as hovering highlight the corresponding positions of molecules across views, enabling detailed tracing of how individual molecules are organized across latent spaces.

LatentFlow is a web-based visual analytics system. 
The Python back-end is powered by Flask\footnote{\url{https://pypi.org/project/Flask/}} and supports operations on embeddings such as clustering and projection.  
The front-end is implemented in TypeScript with React\footnote{\url{https://react.dev/}} and D3.\footnote{\url{https://d3js.org/}} 
All color encodings in the interface are designed to be colorblind-friendly, using distinguishable palettes for categorical and sequential data~\cite{harrower2003colorbrewer}. %~\footnote{\url{https://sronpersonalpages.nl/~pault/}}.
 An online demo is available at \url{https://latent-space-vis.vercel.app/}.

\subsection{Control Panel}
\label{subsec:control_panel}
A compact control panel is positioned at the edge of the workspace and can be expanded on demand (Fig.~\cref{fig:system}G). 
It provides controls for selecting the datasets and data splits (e.g., training or test data) that will be rendered in the interface.
Experts can choose different clustering methods and adjust their parameters, updating the clusters shown in the visualization. These interactions enable examination of how clustering choices affect latent space organization (\textbf{DR1}, \textbf{DR2}).
A preview toggle controls whether an enlarged molecular structure appears on hover, and a size slider adjusts its display size. An export function enables saving bookmarked molecules for further analysis.

\subsection{Modified Sankey View}
\textit{Visual Design.} \emph{LatentFlow} visualizes cluster changes across models using a modified Sankey-style layout, as shown in ~\cref{fig:system}A. 
The layout forms a matrix where rows and columns correspond to two selected dimensions, such as layers and training epochs in \cref{fig:system}A (\textbf{DR1}). 
%This matrix structure allows experts to examine how clusters change along two dimensions in a single view (\textbf{DR1}).
Within each cell (\cref{fig:system}A1), clusters appear as rectangles, with their area encoding the number of molecules within the cluster. 
% Clusters are vertically offset rather than aligned in a single column as in standard Sankey diagrams. 
% This offset creates space for flows to connect clusters from both horizontal and vertical directions.
To connect cells, we draw flows between clusters in both directions that describe how clusters change across models; the width of each flow represents the number of molecules that move from one cluster to the next. Horizontal flows are colored red and vertical ones green, providing better visual separation between the two dimensions.

Cell background colors encode a clustering metric; by default, this metric is the number of clusters in each cell.
We use a sequential color scale to represent increasing values, with darker blue indicating larger numbers of clusters.
%When experts provide labels from the Scatterplot View (\cref{fig:system}C1), metrics such as the adjusted rand index (ARI)~\cite{Hubert1985}, normalized mutual information (NMI)~\cite{strehl2002cluster}, and Fowlkes–Mallows index (FMI)~\cite{Fowlkes01091983} can be used to measure how well the clusters match these labels.
% {\rev{Changes between cells are represented by rectangles in the gaps between rows and columns (\cref{fig:system}A2), colored according to the percentage of molecules that change cluster assignment. As with flows, vertical gaps are shown in green and horizontal gaps in red; darker colors indicate larger values.}}
Changes in cluster membership between cells are represented by gap heatmaps, shown as rectangles in the gaps between rows and columns (\cref{fig:system}A2). Their color encodes the percentage of molecules that change cluster assignment. As with flows, vertical gaps are shown in green and horizontal gaps in red; darker colors indicate larger values.

\textit{Interactions.} 
Experts can toggle the visibility of nodes, paths, cell background colors, and gaps, allowing the viewer to focus on specific elements.
To this end, the Modified Sankey View also supports value -based filtering (\textbf{DR2}) of flows (\cref{fig:system}A3) and cells.
%A link-value filter (\cref{fig:system}A3) shows paths within a selected range of flow values, removing weak transitions. 
%An epoch transition filter removes cells with small changes based on a threshold to focus on states with larger changes.

Experts can select a cell to highlight its row and column, or select a cluster to highlight its flows and cells. %selections are linked across the view.
These interactions enable experts to inspect the molecules within clusters and trace them across latent spaces (\textbf{DR3}).
The axes can be swapped, and the order of models can be reversed.
Pan and zoom interactions support navigation across large matrices as the number of models increases (\textbf{DR2}).

\begin{figure}[t]
    \centering
    \includegraphics[width=0.85\linewidth]{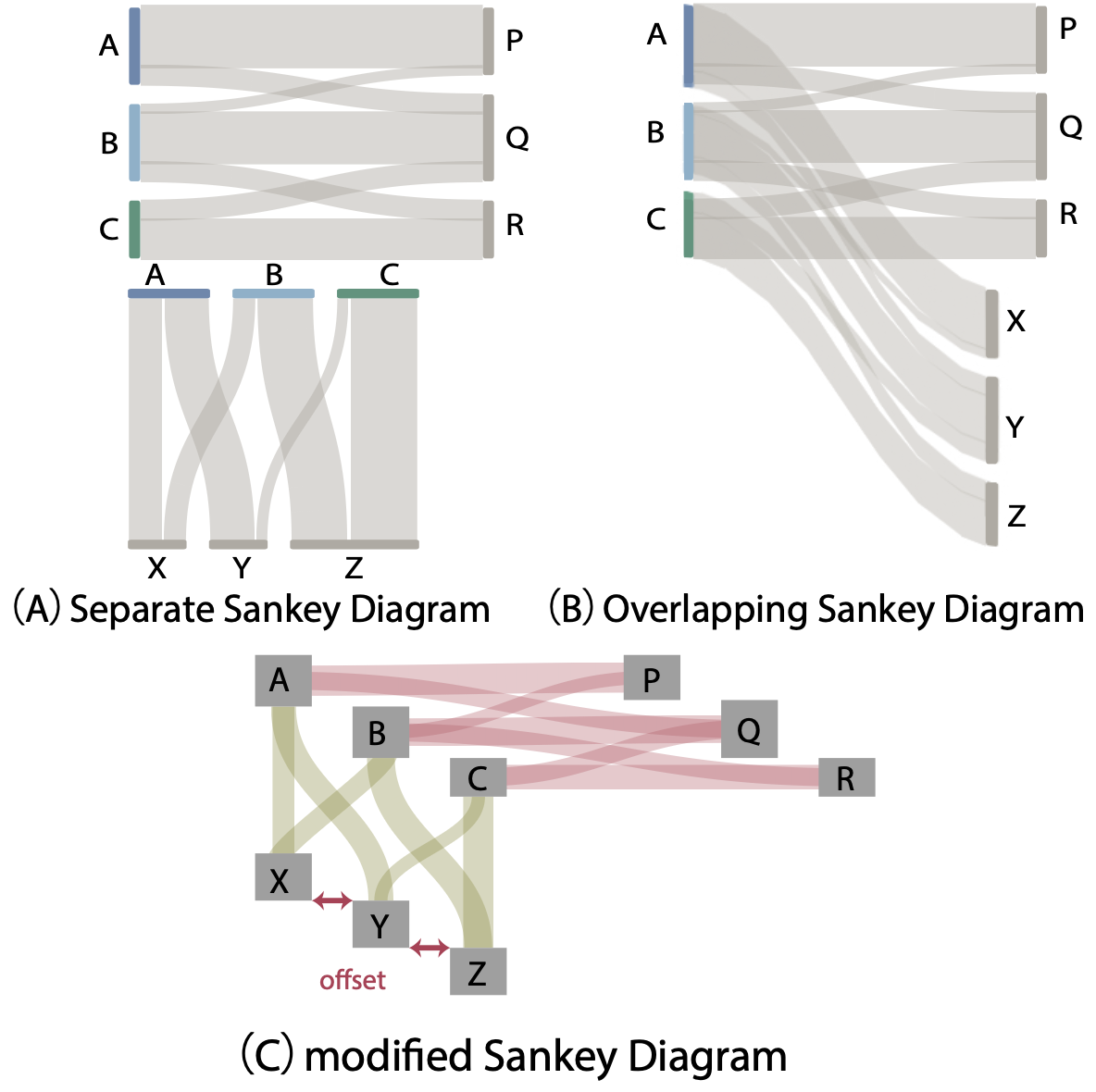}
    \caption{Several designs for the Sankey View; C is employed in \emph{LatentFlow}.}
    \label{fig:alternative}
\end{figure}

\textit{Design Alternatives.}
We considered several alternative designs for visualizing cluster transitions across two dimensions, as shown in \cref{fig:alternative}. 

A straightforward approach is to juxtapose two standard Sankey diagrams (\cref{fig:alternative}A), one arranged horizontally and the other vertically. %, each representing transitions along a single dimension. 
However, this design makes it difficult to establish correspondence between nodes across the two views, requiring users to mentally match clusters. This cognitive burden increases further as the number of layers or epochs grows.
%In addition, as the number of layers or epochs increases, the layout becomes difficult to scale, and the two views cannot be easily aligned.

Another alternative is to overlay two Sankey diagrams (\cref{fig:alternative}B), for example, by rotating one diagram and placing it on top of the other. 
While this design shows both dimensions to be shown in a single view, it introduces severe visual clutter due to overlapping flow paths, an issue again amplified by scale. %As the number of models increases, these overlaps become more pronounced, making it difficult to follow individual transitions.

Our design addresses these limitations by restructuring the layout of Sankey nodes (\cref{fig:alternative}C). 
Instead of aligning nodes in one direction, clusters within each cell are offset to create space for both directions. 
Flows do not occupy the full extent of each node, which reduces overlap and improves readability. 
This design supports the simultaneous visualization of transitions along two dimensions while maintaining scalability and preserving the ability to trace cluster relationships.
\begin{figure*}
    \centering 
    \includegraphics[width=1.0\linewidth]{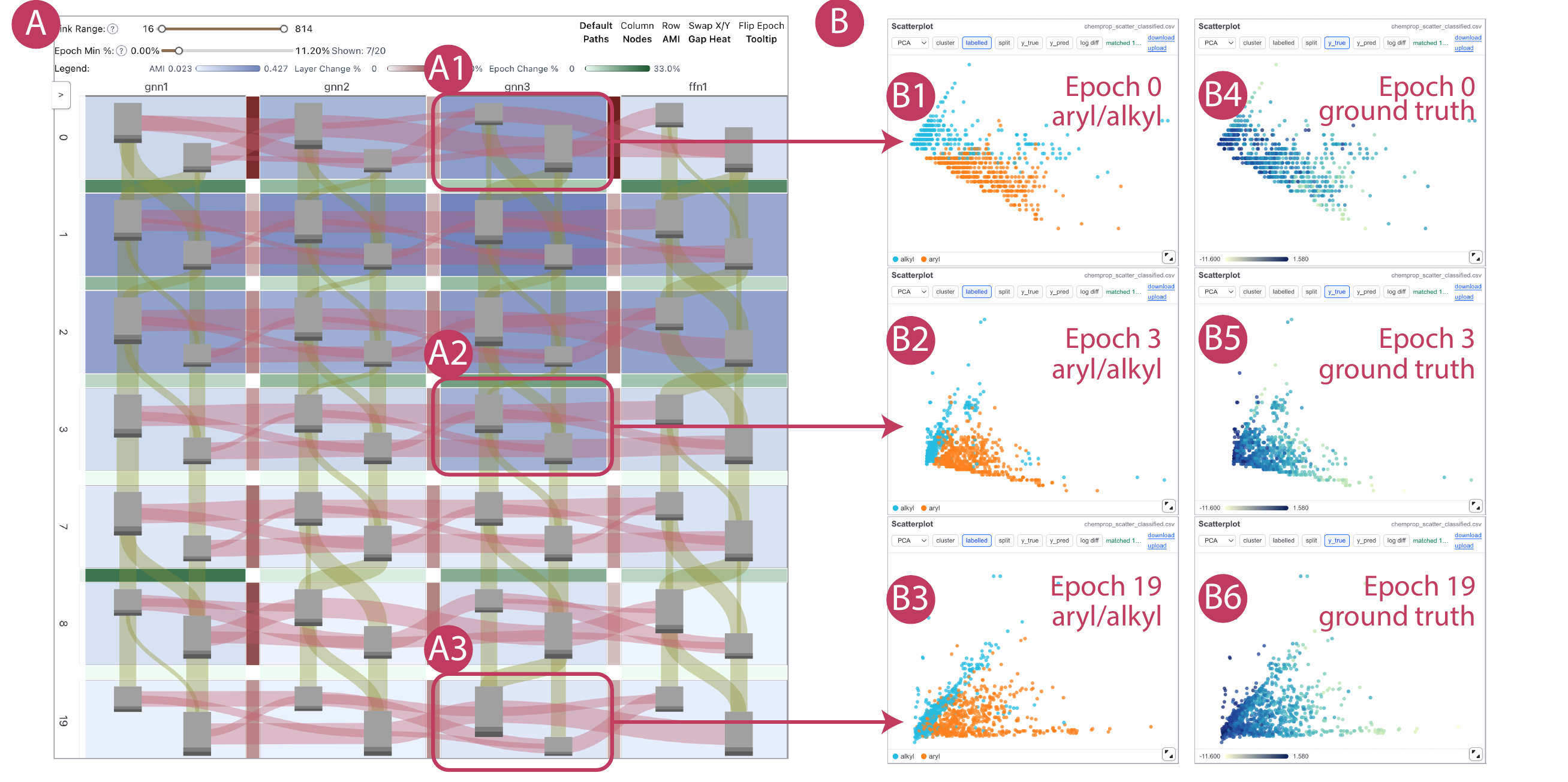}
    \caption{Case study 1: analysis of a Chemprop model showing the transition from taxonomy-aligned to property-driven latent space organization. 
    (A) The Modified Sankey View shows that early epochs exhibit strong agreement with \textit{aryl}/\textit{alkyl} labels (darker AMI), while later epochs diverge as cluster assignments reorganize. 
    (B) Scatterplot Views show that clusters only partially correspond to the \textit{aryl}/\textit{alkyl} taxonomy, with mixed labels within cluster regions. Across epochs (0, 3, 19), label-based separation weakens while a smoother organization aligned with ground-truth solubility emerges. 
    }
    \label{fig:case_chemprop}
\end{figure*}
\subsection{Sankey Views}
Two additional Sankey Views (\cref{fig:system}B) are placed to the right and below the Modified Sankey View to show cluster transitions along a given dimension.
Selecting a cell in the Modified Sankey View opens the detailed Sankey View for the corresponding layer and epoch. 
The horizontal Sankey View shows cluster transitions across layers for the selected epoch (\cref{fig:system}B1), while the vertical Sankey View shows transitions across epochs for the selected layer (\cref{fig:system}B2).

In the Sankey View, rectangles represent clusters, corresponding to the same grouping of molecules as in the Modified Sankey View.
When a cell is selected, clusters across the Modified Sankey View and the detailed Sankey View are colored consistently to facilitate comparison across views. %with different clusters shown in different colors, making them easy to correlate across views.
Each node is shaded according to the split across training and test samples; the darker portion of each node represents test molecules.
Hovering over a molecule in the Scatterplot View, Cluster Member View, or Top-k Panel shows its trajectory (\cref{fig:system}B3) as a path through nodes across model states, providing details on changes not shown in aggregated flows (\textbf{DR2}).

\subsection{Scatterplot View}
\label{subsec:scatterplot}
The Scatterplot View (\cref{fig:system}C) shows molecules projected onto 2D space for cells selected in any Sankey View. 
Each point represents a molecule and is positioned according to its projected coordinates. 
By default, points are colored by their cluster assignment.%, showing how molecules are arranged within each cluster. 
The color scheme can be modified using the scatterplot toolbar (\cref{fig:system}C1), which can encode expert-uploaded labels, the train/test split, ground-truth targets, model predictions, and prediction error. Experts can also upload external labels (\cref{fig:system}C2) to compare clusters with domain-defined groupings (\textbf{DR3}, \textbf{DR4}).
Color encoding and filtering enable comparison of cluster structure with chemical properties and labels (\textbf{DR2}).

The Scatterplot View provides additional interactions.
Pan and zoom support navigation in dense regions. 
Hovering over a point reveals a tooltip showing the molecule's 2D structure (\cref{fig:system}C3), which is rendered from its SMILES representation by RDKit, along with basic metadata (\textbf{DR4}).
Hovering over a point also highlights its trajectory across clusters in the Sankey View, enabling inspection across latent spaces and connecting embeddings to molecular structures (\textbf{DR3}).

\subsection{Cluster Member Panel}
Selecting a cluster in the Modified Sankey View, Sankey View, or Scatterplot View displays its molecules in a separate panel called the Cluster Member Panel (\textbf{DR2, DR3}).
This panel is organized as a paginated table, where each molecule is rendered using RDKit. 
Hovering over a molecule highlights it in the scatterplot as well as its trajectory in the Sankey View, allowing experts to locate the molecule throughout the interface. 
Experts can bookmark molecules, which can later be exported, by clicking the bookmark icon (\cref{fig:system}D1).

\subsection{Top-k Panel}
The Top-$k$ Panel shows cluster-level information in three columns (\cref{fig:system}E), based on the top-$k$ molecules closest to the cluster centroid in the embedding space. 
The value of $k$ can be adjusted through a drop-down menu (\cref{fig:system}E1) to change the number of representative molecules.
The first column (\cref{fig:system}E2) uses ridge plots to show the distribution of model outputs for molecules in the cluster. 
These plots show the distributions of ground-truth values and predictions, enabling the assessment of prediction accuracy and variation within each cluster (\textbf{DR2}).
The second column (\cref{fig:system}E3) shows common substructures extracted from the top-$k$ molecules in the cluster (\textbf{DR4}). 
These substructures are obtained by matching shared subgraphs across molecules and highlight recurring structural patterns. 
The Tanimoto similarity (c.f. \cref{subsec:chembg}) values are also shown to indicate how similar the molecules are.
The third column (\cref{fig:system}E3) shows the top-$k$ molecules.
Hovering over a molecule highlights it in the Scatterplot View and its trajectories in the Sankey View, enabling cross-view comparison (\textbf{DR3}).

\subsection{Structure Editor}
\label{subsec:struc_editor}
A Structure Editor (\cref{fig:system}F) enables the creation of chemical substructures using an embedded molecular editor~\cite{bienfait2013jsme}, a sketching tool commonly used by chemists for drawing molecular structures.
Once a structure is defined, the system searches for molecules containing the substructure and highlights matches across views in the system.
Matches are highlighted in the Scatterplot View and are listed in the Cluster Member Panel.
These interactions enable comparison of user-defined substructures with clusters and latent spaces (\textbf{DR4}).

\begin{figure}[t]
    \centering
    \includegraphics[width=0.9\linewidth]{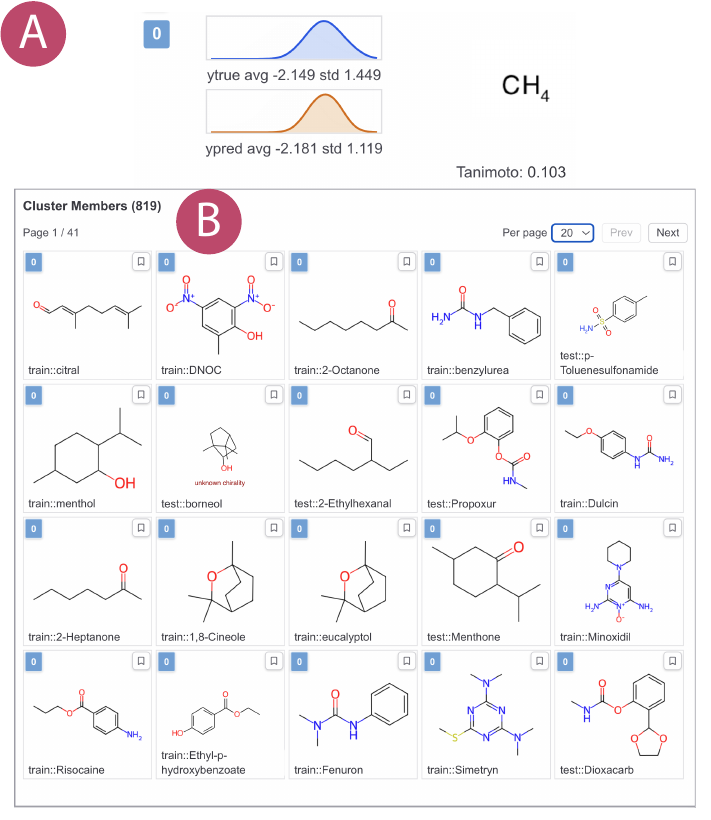}
    \caption{Case study 1: Cluster 0 at \texttt{Epoch 19/gnn3}
    (A) The Top-$k$ Panel shows representative molecules and shared substructures. 
    (B) The Cluster Member Panel shows individual molecules in the cluster. }
    \label{fig:case_chemprop2}
\end{figure}

\section{Case Studies}
\label{sec:case_studies}
% To evaluate the effectiveness of \emph{LatentFlow}, we present two case studies conducted with a domain expert.  
To evaluate the effectiveness of \emph{LatentFlow}, we present two case studies conducted with the same domain expert who participated in the design process.
Throughout the five-month development period, the expert independently used \emph{LatentFlow} to explore multiple datasets and models during regular analysis sessions.
The findings reported in the following case studies reflect observations made during these exploratory analyses.
In Case Study 1, we analyze a Chemprop model~\cite{chemprop} using user-defined labels to examine how latent spaces evolve from coarse taxonomy-based groupings to property-driven organization. 
In Case Study 2, we compare two DimeNet models~\cite{doi:10.26434/chemrxiv.10001648/v1} trained with and without conformer augmentation to examine how data augmentation affects latent space organization and cluster structure.
Together, these case studies demonstrate how \emph{LatentFlow} supports tracing latent space changes, inspecting cluster structure, and relating patterns to molecular properties.
%These studies demonstrate how the system supports analysis of latent spaces from two complementary perspectives: comparing how different models organize the same data and examining how a single model refines its representations across layers. 

\subsection{Case Study 1: From Taxonomy to Property-Driven Clustering}

\textbf{Overview analysis with external labels (DR2, DR3).}
This case study focuses on a 2D graph neural network, Chemprop, trained on the MoleculeNet ESOL dataset~\cite{wuMoleculeNetBenchmarkMolecular2018} to predict the solubility of small organic molecules in water.
The expert begins with a coarse taxonomy that labels molecules as \textit{aryl} or \textit{alkyl}. 
\textit{Aryl} molecules are dominated by aromatic ring motifs such as benzene rings~\includegraphics[height=1.2em]{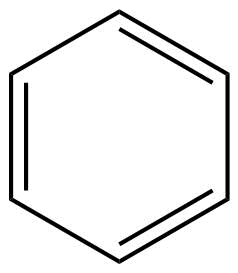}, while \textit{alkyl} molecules are primarily chain-like~\includegraphics[height=0.8em]{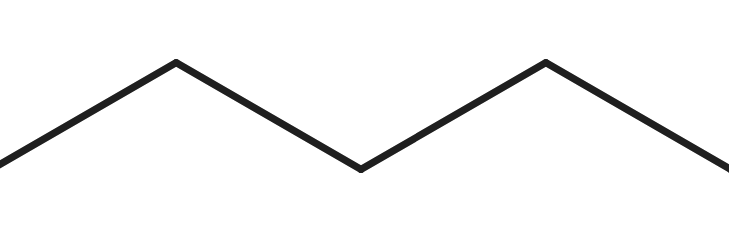}. 
Although this taxonomy captures a broad structural distinction, it does not directly determine solubility, so it serves as a reference rather than as the target explanation.

To compare the latent space to this reference, the expert uploads the \textit{aryl} / \textit{alkyl} labels and encodes AMI (\cref{subsec:overview}) as the background color in the Modified Sankey View, as shown in \cref{fig:case_chemprop}A. 
% Darker cells indicate stronger agreement between the current clusters and the labels. 
The cells are shown with a blue background, where darker blue indicates higher AMI values and stronger agreement between the current clusters and the labels.

\begin{figure}[t]
    \centering
    \includegraphics[width=0.9\linewidth]{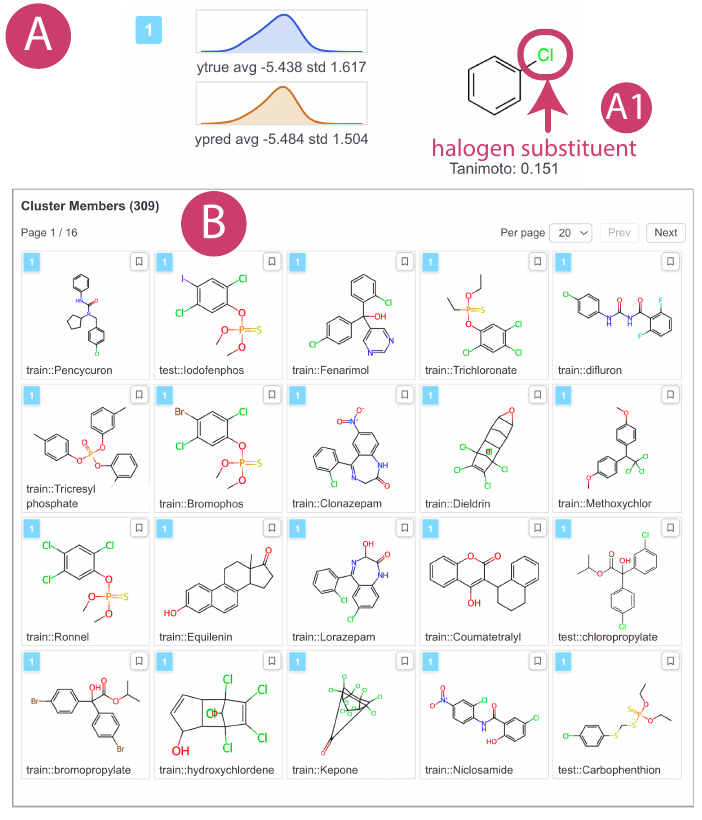}
    \caption{Case study 1: Cluster 1 at \texttt{Epoch 19/gnn3}
    (A) The Top-$k$ Panel shows representative molecules and shared substructures. 
    (B) The Cluster Member Panel shows individual molecules in the cluster. }
    \label{fig:case_chemprop3}
\end{figure}

The expert first examines the Modified Sankey View to assess how the latent space aligns with the uploaded labels across model states. 
% Across the entire view, early epochs appear consistently darker, while later epochs become lighter (\cref{fig:case_chemprop}A), indicating that the model’s organization is initially aligned with the taxonomy but gradually diverges from it. 
Across the view, early epochs appear darker, while later epochs become lighter (\cref{fig:case_chemprop}A). Darker blue indicates higher AMI values, showing that clusters in early epochs closely match the \textit{aryl}/\textit{alkyl} labels, whereas lower AMI values in later epochs indicate that clusters no longer correspond to this taxonomy.
Flow patterns indicate that most reorganization occurs in early epochs, with larger flows reflecting substantial changes in cluster assignment, while later epochs show smaller flows, indicating that the latent space becomes more stable over time.

\textbf{Track label alignment across model states (DR1, DR3).}
To examine this transition in detail, the expert focuses on the \texttt{gnn3} layer, where this pattern is most evident, and selects \texttt{Epoch 0} (\cref{fig:case_chemprop}A1), \texttt{Epoch 3} (\cref{fig:case_chemprop}A2), and \texttt{Epoch 19} (\cref{fig:case_chemprop}A3) for further inspection.

In the Scatterplot View (\cref{fig:case_chemprop}B), coloring points by \textit{aryl} / \textit{alkyl}, shows a clear separation at \texttt{Epoch 0}, with orange (\textit{aryl})  and blue (\textit{alkyl}) points occupying distinct regions of the projection (\cref{fig:case_chemprop}B1). 
By \texttt{Epoch 3}, the two colors begin to intermix within the same regions, appearing together in overlapping areas rather than forming distinct groups (\cref{fig:case_chemprop}B2). 
By \texttt{Epoch 19}, the two colors are fully mixed across the projection, and no clear boundary remains between the two label groups (\cref{fig:case_chemprop}B3). 
This progression indicates that the latent space gradually moves away from the initial taxonomy-based organization.

However, when recoloring the points by ground-truth solubility, a different pattern appears.
Here, a sequential blue color scale encodes solubility, with darker shades indicating higher values and lighter shades lower values. 
In early epochs (\cref{fig:case_chemprop}B4--B6), the distribution appears irregular, with no clear spatial trend. 
As training progresses, the structure becomes more pronounced. 
By \texttt{Epoch 19} (\cref{fig:case_chemprop}B6), a clear color gradient emerges, with darker blue points (higher solubility) concentrated on one side and lighter blue points (lower solubility) concentrated on the other. 
This contrast between early and later epochs indicates that the latent space reorganizes according to a property more directly related to the prediction task.

\begin{figure*}
    \centering
    \includegraphics[width=0.9\linewidth]{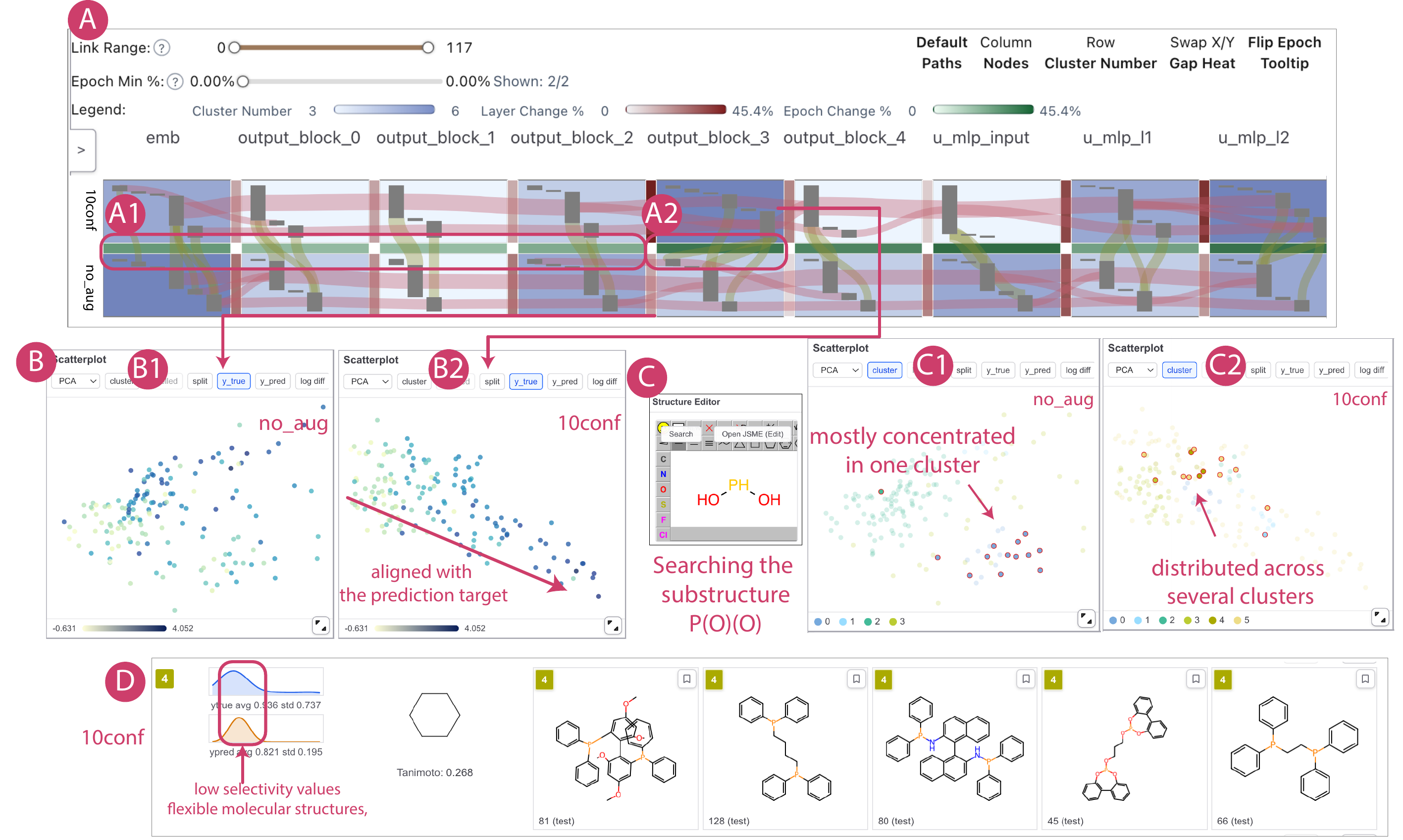}
    \caption{Case study 2: comparison of two DimeNet models with the same architecture but different training inputs. (A) The Modified Sankey View shows that, starting from \texttt{output\_block\_3}, cluster assignments diverge between the two models (\texttt{10conf} model in the top row and \texttt{noaug} model in the bottom row). (B) \emph{Scatterplot Views} at the selected layer show that the \texttt{10conf} model produces clusters that are better aligned with target values, whereas the \texttt{noaug} model remains more heterogeneous. (C) Substructure queries for the phosphite substructure \texttt{P(O)(O)} from SMILES show that molecules sharing the same motif are grouped into a single cluster in the \texttt{noaug} model but distributed across multiple clusters in the \texttt{10conf} model. (D) In the final layer, clusters correspond to distinct structural patterns and levels of prediction accuracy.}
    \label{fig:case_dimenet}
\end{figure*}

\textbf{Relate cluster structure to molecular features (DR3, DR4).}
To characterize these clusters, the expert first examines the Top-$k$ Panel (\cref{fig:case_chemprop2}, \cref{fig:case_chemprop3}). 
The panel shows that Cluster 0, associated with higher predicted solubility, contains methane-like substructures ($\mathrm{CH_4}$) (\cref{fig:case_chemprop2}A). 
In contrast, Cluster 1, associated with lower predicted solubility, contains aromatic ring substructures and a halogen substituent such as $\mathrm{Cl}$ (\cref{fig:case_chemprop3}A1). 
At this level, the extracted substructures show that they largely reproduce the initial \textit{aryl}/\textit{alkyl} distinction, with Cluster 0 appearing more alkyl-like (containing ~\includegraphics[height=0.8em]{myfigs/alkyl.png}) and Cluster 1 more aryl-like (containing ~\includegraphics[height=1.2em]{myfigs/benzene.png}). 
However, the extracted substructures do not explain the difference in solubility between the clusters

\textbf{Inspect molecules to identify chemical factors (DR3, DR4).}
To examine this further, the expert opens the Cluster Member Panel and inspects individual molecules within each cluster. 
In Cluster 0, the molecules are generally small and structurally simple, with few substituents (\cref{fig:case_chemprop2}B). 
In contrast, molecules in Cluster 1 are larger and more complex, and consistently contain halogen atoms such as $\mathrm{Cl}$  (\cref{fig:case_chemprop3}B). 
This difference is visible across multiple molecules in each cluster, rather than being driven by a few outliers. 
These observations indicate that molecular size and halogenation contribute to the separation between clusters.

Returning to the Top-$k$ Panel, the expert verifies that this pattern is already reflected in the extracted substructures. 
The presence of $\mathrm{Cl}$ appears as a recurring feature in Cluster 1, confirming that the cluster-level summary captures this signal as shown in \cref{fig:case_chemprop3}A1. 
Together, these observations indicate that the model organizes molecules based on factors beyond the coarse \textit{aryl} / \textit{alkyl} taxonomy, incorporating additional structural properties such as size and halogenation.

\textbf{Takeaway.}
This case study shows that latent space organization evolves from coarse taxonomy-based groupings to property-driven structure during training. 
While early representations align with the \textit{aryl} / \textit{alkyl} distinction, this alignment weakens as the model reorganizes molecules according to factors more directly related to solubility. 
Through cluster- and molecule-level inspection, the expert identifies that molecular size and halogenation, rather than the coarse taxonomy alone, explain the observed separation. 
This analysis follows the multi-level design of \emph{LatentFlow}. 
The expert first uses the Modified Sankey View to identify when the latent space diverges from the initial taxonomy, then uses the Scatterplot View to examine how spatial organization changes, followed by the Top-$k$ Panel to relate clusters to structural patterns, and finally the Cluster Member Panel to verify these patterns at the level of individual molecules. 
This stepwise workflow enables the expert to move from a high-level overview to a detailed, property-driven interpretation of latent space organization.

\subsection{Case Study 2: Effect of Conformer Augmentation on Latent Space Organization}
In this case study, the expert analyzes two 3D graph neural network models, trained to predict experimental catalyst selectivity. 
The baseline model (\texttt{noaug}) is trained using a single structure for each molecule, and
the augmented model (\texttt{10conf}) is trained on multiple conformations per molecule, thereby using data augmentation to expose the model to a broader set of valid 3D geometries for the same molecular graph.
%Here, a conformation refers to a distinct 3D arrangement of the same atoms, arising from rotation about rotatable bonds. 
In practice, molecules naturally sample multiple conformations due to thermal motion, and these conformations can exhibit slightly different properties, with observed behavior reflecting their ensemble.
This data augmentation resulted in lower error and better extrapolation to out-of-distribution catalysts~\cite{gomez2018automatic}, but \emph{post hoc} performance metrics do not explain how data augmentation impacts the models' latent spaces.
Our expert aimed to examine how data augmentation affects the organization of the latent spaces for the Dimenet architecture.
The expert wants to understand whether conformer augmentation leads to more structured and chemically meaningful groupings of molecules compared to the baseline model. To do this, the expert used \emph{LatentFlow} to compare the latent spaces of all layers across all training epochs of the two models. 

\textbf{Locate divergence in latent spaces (DR1, DR2).}
In the Modified Sankey View (\cref{fig:case_dimenet}A), the expert compares the two models across layers by examining the gap heatmaps, where the shade of the green color encodes the fraction of molecules that change cluster assignment between corresponding cells. 
In the early layers (\texttt{output\_block\_0} to \texttt{output\_block\_2}), the gaps are shown in light green (\cref{fig:case_dimenet}A1), indicating that only a small fraction of molecules changes clusters and that the two models group molecules similarly. 
At \texttt{output\_block\_3}, the gap color becomes darker green (\cref{fig:case_dimenet}A2), indicating that a larger fraction of molecules are reassigned to different clusters. 
This change shows that the cluster structure begins to diverge between the two models. 
Based on this observation, the expert selects \texttt{output\_block\_3} for further inspection, as it marks the first layer where the latent spaces differ.

\textbf{Compare cluster separation (DR2).}
\Cref{fig:case_dimenet}B shows the \emph{Scatterplot Views} for \texttt{output\_block\_3} of the two models.
In the \texttt{noaug} model (\cref{fig:case_dimenet}B1), molecules form several clusters, but when colored by ground-truth values, points of different colors are mixed within the same cluster regions. 
High- and low-value points appear together across the projection, with no consistent spatial ordering, indicating that spatial proximity does not correspond to similarity in the target.
In contrast, the \texttt{10conf} model (\cref{fig:case_dimenet}B2) shows a more structured pattern. 
When colored by ground-truth values, the colors are arranged along a clear left-to-right gradient, with similar colors grouped together and values changing progressively across the projection. 
Points with similar target values are located close to each other, while points with different values appear in different regions.
These differences indicate that \texttt{10conf} organizes molecules such that spatial position reflects target similarity, whereas \texttt{noaug} shows no consistent relationship between position and target values.

\textbf{Test substructure-driven grouping (DR4).}
To test whether latent-space clustering captures groups of catalysts with similar structural features, the expert queries a phosphite substructure, $\mathrm{P(O)(O)}$, using the Structure Editor (\cref{fig:case_dimenet}C). 
In \texttt{noaug}, the highlighted molecules are almost entirely assigned to the same cluster, as indicated by their uniform light-blue color corresponding to Cluster 1 (\cref{fig:case_dimenet}C1). 
This shows that most matching molecules are grouped into a single cluster, suggesting that the model organizes these data primarily according to a shared molecular motif. 
In contrast, in \texttt{10conf}, the highlighted molecules appear in multiple colors (\cref{fig:case_dimenet}C2), reflecting assignment to different clusters (e.g., Cluster 1, 4, and 5). 
The same substructure is distributed across several clusters rather than grouped into one, indicating that \texttt{10conf} is less dominated by a single substructure and instead captures additional differences among these data points.

\textbf{Compare cluster structures (DR3, DR4).}
The expert continued tracing \texttt{10conf} through later layers. 
The Modified Sankey View (\cref{fig:case_dimenet}A) shows that further reorganization occurs before the final layers stabilize. 
Therefore, the expert examined the final layer to understand the resulting grouping structure. 
Inspecting the final layer of \texttt{10conf} reveals that it learns a latent-space organization that more effectively captures structure–selectivity relationships by grouping molecules with related structural features and, consequently, similar predicted selectivities. 
For example, Cluster 4 (\cref{fig:case_dimenet}D) corresponds to catalysts with low selectivity values and relatively flexible molecular structures, features consistent with their reduced selectivity.
This suggests that augmentation improves the model’s ability to learn latent representations that better reflect meaningful structure–selectivity relationships.

\textbf{Takeaway.}
This case study shows how \emph{LatentFlow} supports a stepwise, linked analysis of latent-space organization across complementary views.
The Modified Sankey View helps identify stages with substantial reorganization, the Scatterplot View reveals how molecules are partitioned, and the substructure and Top-$k$ Panel help interpret the structural features associated with each group.
By linking these views, the expert can seamlessly shift from identifying where changes occur to understanding how those changes manifest in the learned representation and what molecular features underlie them.

\section{Discussion and Conclusions}
In this work, we introduced \emph{LatentFlow}, a visual analytics system for helping computational chemists investigate how molecular structures and properties are organized in latent spaces to reveal chemically meaningful patterns.
\emph{LatentFlow} was designed to address three challenges faced by computational chemists when interpreting molecular GNN latent spaces: tracking how molecular representations change across model states; identifying meaningful structural patterns; and relating these representations to chemical knowledge.
By linking latent-space organization to molecular structures and properties, the system helps reveal structure--property relationships and chemical factors underlying model predictions.
We hope that \emph{LatentFlow} provides a foundation for future visual analytics tools that support the interpretation, diagnosis, and development of embedding-based machine learning models in chemistry and related scientific domains.
% Although \emph{LatentFlow} was developed for molecular GNNs, its workflow is broadly applicable to other settings that produce evolving embeddings. 
% We therefore view \emph{LatentFlow} as a general approach for inspecting changing representations in machine learning models, rather than as a visualization tied to a single chemistry task.

\subsection{Lessons Learned}
One lesson from our design process is that latent space analysis is most useful when the expert can start from a coarse overview and then move to concrete molecules~\cite{Keim2008}. 
In practice, the expert first uses the Modified Sankey View to find cells with significant changes in cluster membership. %large changes, stable stages, or transitions that are driven by a small number of molecules. 
Once transitions of interest are identified, the expert then uses the Scatterplot View, Cluster Member View, and Top-$k$ Panel to inspect the molecules behind those transitions. 
These views are highly complementary, with the Sankey View showing \emph{where} significant transitions happen (i.e., across either layers or time), while the detailed molecule-level views show \emph{what} the transitions contain.
%This workflow was more useful than looking at one view alone, because the Sankey View show where changes happen, while the molecule-level views show what those changes contain.

We also found that introducing external domain-specific knowledge was valuable when examining latent spaces. %is most valuable when it can be compared directly with model outputs. 
The Structure Editor enables experts to draw substructures and immediately search for matching molecules across views, supporting hypothesis testing without switching context or relying on external programs. %the system. 
%Instead of treating chemical knowledge as a separate annotation step, \emph{LatentFlow} makes it part of the analysis loop. 
%This makes it easier for experts to check whether a model organizes molecules in a way that matches known substructures or prediction patterns.

\subsection{Generalization to Other Models and Domains}
Although LatentFlow was developed in close collaboration with a single domain expert, the challenges that motivated the system were not tied to a specific project or dataset. During the design process, the expert repeatedly emphasized the difficulty of understanding how molecular representations evolve during training, identifying when chemically meaningful structures emerge, and relating representation changes to prediction behavior. These challenges arise naturally in workflows that rely on learned molecular embeddings. While our study does not claim that these needs are representative of all molecular machine learning practitioners, they motivated a set of analytical tasks that may also be relevant in other embedding-based molecular analysis scenarios.

\subsection{Limitations and Future Work}
\emph{LatentFlow} currently shows only two model-state dimensions at a time in the Modified Sankey View. 
This approach works well for the cases we studied, such as layer and epoch or model configurations, but it does not display all possible dimensions together. 
When domain experts want to compare another factor, such as a different model configuration or parameter setting, they must switch the axes. 
Future work could explore how the Sankey View could be extended to higher dimensions. %among multiple state dimensions without requiring repeated switching.

The insights gleaned from the system are highly dependent on choosing suitable clustering and projections.
Different clustering methods can produce different groupings, and different projection methods can change how molecules appear in the Scatterplot View.
Although \emph{LatentFlow} employs temporally consistent projection strategies to support comparison across model states, low-dimensional projections inevitably introduce distortions of the underlying high-dimensional structure. Different projection methods may emphasize different aspects of the data, leading to different visual organizations of the same latent space.
We made these choices configurable because no single setting works best for every dataset.
However, this flexibility requires comparison of multiple views and parameter settings to determine whether an observed pattern is robust across different projections and clustering methods.
Future work could rank or recommend settings based on how consistently they preserve structural patterns and cluster transitions.

A related limitation is that some clustering methods can produce a large number of clusters.
The number of clusters typically ranged up to 10, for which the visualization remained easy to interpret.
As the number of clusters increases, the Modified Sankey View becomes crowded, increasing visual density and making individual flows harder to distinguish and transitions more difficult to trace.
Although background coloring and gap-based encodings improve readability, they are not sufficient when many clusters are present.
Visual clutter becomes increasingly noticeable once the number of clusters reaches several dozen, particularly when many transitions are present between adjacent model states.
A similar issue arises when experts introduce taxonomies with many label categories.
While the current system supports comparison between clusters and external labels, increasing the number of categories can make color encodings more difficult to interpret and reduce the effectiveness of visual comparison.
Larger taxonomies would likely require category aggregation, hierarchical groupings, or alternative visual encodings.
Future work could introduce additional filtering and simplification mechanisms, such as bundling nearby flows, aggregating small clusters, or supporting hierarchical label structures, to reduce visual complexity while preserving key transition patterns.

Another limitation is that summary views can hide important details. 
Background colors, gap heatmaps, and flow widths highlight main changes, but they may obscure fine-grained variations due to individual molecules that behave differently. 
The Top-$k$ Panel has a similar limitation, as it only shows representative molecules, and it may miss rare cases or small transitions. 
While these summaries improve readability, they can overlook exceptions. 
Future work could add mechanisms to surface these exceptions more directly, as well as improved summarization techniques to better balance overview and detail.

Finally, the current system focuses on inspection and comparison rather than automated explanation. 
\emph{LatentFlow} helps experts identify where changes happen and which molecules are involved, but it does not automatically decide which transitions are relevant. %important. 
In future work, we plan to explore ranking strategies that highlight transitions based on change magnitude, cluster stability, or chemical novelty. 
This would reduce the need for manual examination and help experts locate meaningful patterns more quickly. %move more quickly from a large latent space to the most informative molecular patterns.

\section{Acknowledgments}
This work was supported by the Laboratory Directed Research and Development (LDRD) Program of Lawrence Berkeley National Laboratory and by the U.S. Department of Energy, Office of Science, Advanced Scientific Computing Research (ASCR) program under Contract No. DE-AC02-05CH11231 to Lawrence Berkeley National Laboratory and Award No. DE-SC0023328 to Arizona State University (“Visualizing High-dimensional Functions in Scientific Machine Learning”). 
We would also like to acknowledge the DOE Competitive Portfolios grant and the DOE SciGPT grant.
This research used resources at the National Energy Research Scientific Computing Center (NERSC), a U.S. Department of Energy, Office of Science, User Facility under NERSC Award Number ASCR-ERCAP0026937. We thank the NIH for funding this work at UC Berkeley (1R35GM130387). N.H. was supported by the National Science Foundation Graduate Research Fellowship Program under Grant No. (DGE 2146752). Any opinions, findings, and conclusions or recommendations expressed in this material are those of the author(s) and do not necessarily reflect the views of the National Science Foundation. 

\bibliographystyle{abbrv-doi-hyperref}
\bibliography{ref.bib}
\end{document}